\documentclass{article}
\usepackage{spconf,amsmath,graphicx}
\usepackage{subcaption}
\usepackage{multirow}
\usepackage{caption}
\usepackage{verbatim}
\usepackage{fancyhdr}
\newcommand{\etal}{\textit{et al.}}

\title{Deep Fence Estimation using Stereo Guidance and Adversarial Learning}
\name{Paritosh Mittal, Shankar M Venkatesan, Viswanath Veera, Aloknath De}
\address{Samsung Research Institute, Bangalore}
\begin{document}
\ninept

\maketitle
\begin{abstract}
People capture memorable images of events and exhibits that are often occluded by a wire mesh loosely termed as fence. Recent works in removing fence have limited performance due to the difficulty in initial fence segmentation. This work aims to accurately segment fence using a novel fence guidance mask (FM) generated from stereo image pair. This binary guidance mask contains deterministic cues about the structure of fence and is given as additional input to the deep fence estimation model. We also introduce a directional connectivity loss ($DCL$), which is used alongside adversarial loss to precisely detect thin wires. Experimental results obtained on real world scenarios demonstrate the superiority of proposed method over state-of-the-art techniques.
%Recent works in removing fence rely on capturing a video of an almost static scene.
%They however have 
\end{abstract}
\section{Introduction}
\label{sec:intro}
With the advancements in sophisticated scene capturing devices, images have emerged as one of the most prominent ways of sharing and preserving experiences. Images taken in places like zoos, parks, stadiums etc., often contain an obstruction called fence. It degrades the image by dominating the scene, obstructing important objects and reducing the aesthetic appeal of the image. Hence, they are highly undesirable among users. The problem of fence removal has two main components: (i) fence mask estimation and (ii) in-painting of regions that were originally occluded \cite{Criminisi:2004:RFO:2319036.2320602,Bertalmio:2000:II:344779.344972,yu2018generative,papafitsoros2013combined}. Several popular methods rely on capturing a video \cite{yi2016automatic,Mu2014VideoD,du2018accurate} for fence removal. However, it is difficult for users to carefully record long videos in slow motion with almost static background. Single image based fence estimation methods \cite{jonna2016deep,hettiarachchi2014fence,4815261} do not work well in situations when fence is very thin, out of focus, or when background is very complex and rich in texture.

\begin{figure}
	\label{fig:1}
	\centering
	\begin{subfigure}{0.14\textwidth}
		\includegraphics[width=\linewidth,height=1.8cm]{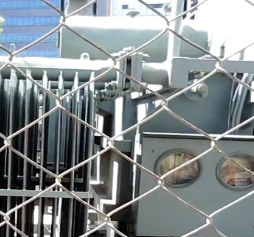}
		\caption{Image}
		\label{fig:1_InpImg} 
	\end{subfigure}
	\begin{subfigure}{0.14\textwidth}
		\includegraphics[width=\linewidth,height=1.8cm]{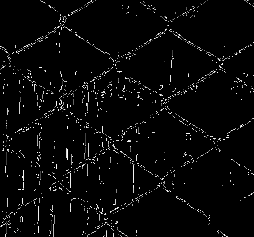}
		\caption{FM}
		\label{fig:1_guidanceMask} 
	\end{subfigure}
	\begin{subfigure}{0.14\textwidth}
		\includegraphics[width=\linewidth,height=1.8cm]{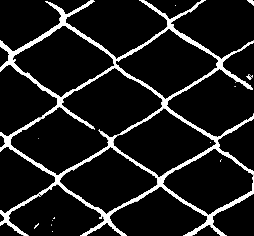}
		\caption{Final Output}
		\label{fig:1_outputImg} 
	\end{subfigure}
	\caption{(c) is the output of DefenceGAN-4c with \textit{DCL}}
	\vspace{-1\baselineskip}
\end{figure}

We introduce a novel method of using stereo images to accurately predict the fence mask. Current smartphones can capture stereo images of a scene aligned in one direction with single user click. We use these images to automatically generate additional assistance which we define as guidance mask (Fig. \ref{fig:1_guidanceMask}). This guidance mask contains the approximate structure of the fence and is very similar to manual drawing/scribbling over fence regions in the image. This strengthens the intuition behind proposing stereo-hardware guidance to perfectly detect fence in difficult situations with complicated and content-rich backgrounds, bad lighting, thin and complex fence shapes etc. The proposed method of guidance generation works best when  the fence plane and camera plane are almost parallel. For the scope of this work, we define fence as a wire mesh and not as a wooden slat. This is because wire mesh like fences are more common in public places like zoos, parks, stadiums etc.

This work introduces two deep learning based fence mask estimation models: (i) \textbf{DefenceGAN-3c}, which uses single image as input and (ii) \textbf{DefenceGAN-4c}, which uses a combination of single image and guidance mask as input. To the best of our knowledge, this is the first work to analyze the role of adversarial loss in refined fence segmentation. We also introduce a novel Directional Connectivity Loss (DCL) which penalizes predicted fence pixels that are not connected to one another. This improves the model's ability to detect thin fence wires. The proposed method is able to take benefit of both: (i) video sequence based methods, by creating a parallax driven guidance mask and (ii) single image based methods, by leveraging deep learning based fence segmentation. The paper is structured as follows. Section \ref{sec:relatedWork} looks at the related work. In Section \ref{sec:propApp},we describe an automatic way of generating guidance mask, modal architecture and the DCL. In Section \ref{sec:Results} the experiments and associated results are detailed. Finally, Section \ref{sec:Conclusion} concludes the paper.
\section{Related Work}
\label{sec:relatedWork}
Major works on removing fence depends on video of any particular scene. It is often assumed that fence is the object closest to camera. Hence parallax for fence pixels are largest between two frames. Xue \etal \cite{xue2015computational} and Yadong Mu \etal \cite{Mu2014VideoD} use this to identify fence but fail to produce intended results when background objects are non-static or in close proximity with fence. Yi \etal \cite{yi2016automatic} attempts to group pixels based on color and motion using graph-cut optimization and spatio-temporal refinement across multiple frames. This approach works well with dynamic background. Du \etal \cite{du2018accurate} use semantic segmentation on single frame to predict approximate mask. They perform temporal refinement across multiple predicted masks from a video. The initial mask eliminates background and eases refinement. 
%  The approach works well with dynamic background but produces imprecise fence mask when foreground and background color are similar.
However, it is difficult for users to capture videos of every scene. Hettiarachchi \etal \cite{hettiarachchi2014fence} proposed a single image based method to detect fence in fourier domain. Their approach aimed at leveraging the quasi-periodic texture of fence but required manual thresholding to eliminate background. Jonna \etal \cite{jonna2016deep} used deep learning to detect fence joints from single images and joined them using straight lines. But fence wires are never truly straight. Jonna \etal \cite{jonna2017stereo} made an attempt to estimate fence using a stereo image pair. They use morphological transformations on disparity maps to compute fence mask. The method did not utilize the structure of fence and worked well with considerable background-foreground separation. We exploit the learned features alongside a parallax driven real-time guidance mask for refined prediction. The stereo guidance mask is computed using the already available multi-camera setup with a single user click thus obviating the problems due to dynamic backgrounds.

Fence segmentation from a given image can be described as an image translation task. The seminal works of Isola \etal \cite{pix2pix2016} and Zhu \etal \cite{CycleGAN2017} use adversarial learning to generate realistic output that are conditioned on input images. Pioneering works of Xue \etal \cite{Xue2018} and Majurski \etal \cite{Majurski_2019_CVPR_Workshops} on segmentation of cells in medical imaging with complex background and inconsistent cell shapes have also leveraged adversarial learning for refined and generalized segmentation. Since fence wires are also inconsistent in shape and difficult to detect, we use an adversarial loss to refine our predictions.
\section{Proposed Approach}
\label{sec:propApp}
Consider a stereo image pair with images left (L) and right (R). In principle, if we know the amount of foreground parallax ($s$) for fence pixels then simply shifting the image (L) by $s$ pixels and subtracting it with the stereo counterpart (R) can eliminate the fence. But a lot of background information is also lost due to \textbf{accidental subtractions}. We define it as the phenomenon when two pixels, each from L and R, match closely in value and get subtracted even when they do not share the given parallax. For example the pixels from a stereo image pair with uniform planar blue sky will always match irrespective of the actual pixel shift. Regions highlighted in Fig. \ref{fig: 2}a \& d illustrate major accidental subtractions. The process of subtraction can still provide critical information as guidance (FM). Section \ref{subsec: stereo-guidance} explains a way of generating FM with minimal accidental subtractions. 
\subsection{Stereo Guidance Mask Generation}
\label{subsec: stereo-guidance}
To avoid accidental subtractions, we compute canny edges \cite{canny1987computational} of stereo frames ($\sim88\%$ $\downarrow$). The canny edges ($CL$ \& $CR$) are binary. We generate FM by performing dual subtraction as explained in Eq. \ref{eq:1}. This gets back only the edges that match for a given pixel shift $i$.
% with $0$ signifying background and $1$ signifying prominent edge pixel. 
%Computation using $CL$ and $CR$ results in an $\sim88\%$ drop in number of accidental subtractions. 
%We leverage the idea of subtraction between stereo frames to generate an approximate Fence Mask ($FM$). perform
\begin{equation} \label{eq:1}
	\vspace{-0.25\baselineskip}
	\begin{split}
		FM_{i} &= f(CR - f(CR-CL_{i}))\text{ $where,$ }\\ f(x) &= 
		\begin{cases}
			x & \text{if $x>0$}.\\
			0 & \text{if $x \leq 0$}.
		\end{cases}
	\end{split}
	\vspace{-0.5\baselineskip}
\end{equation}
%\vspace{-0.25\baselineskip}
\begin{figure}
	\centering
	\includegraphics[width=0.8\linewidth,height=3.3cm]{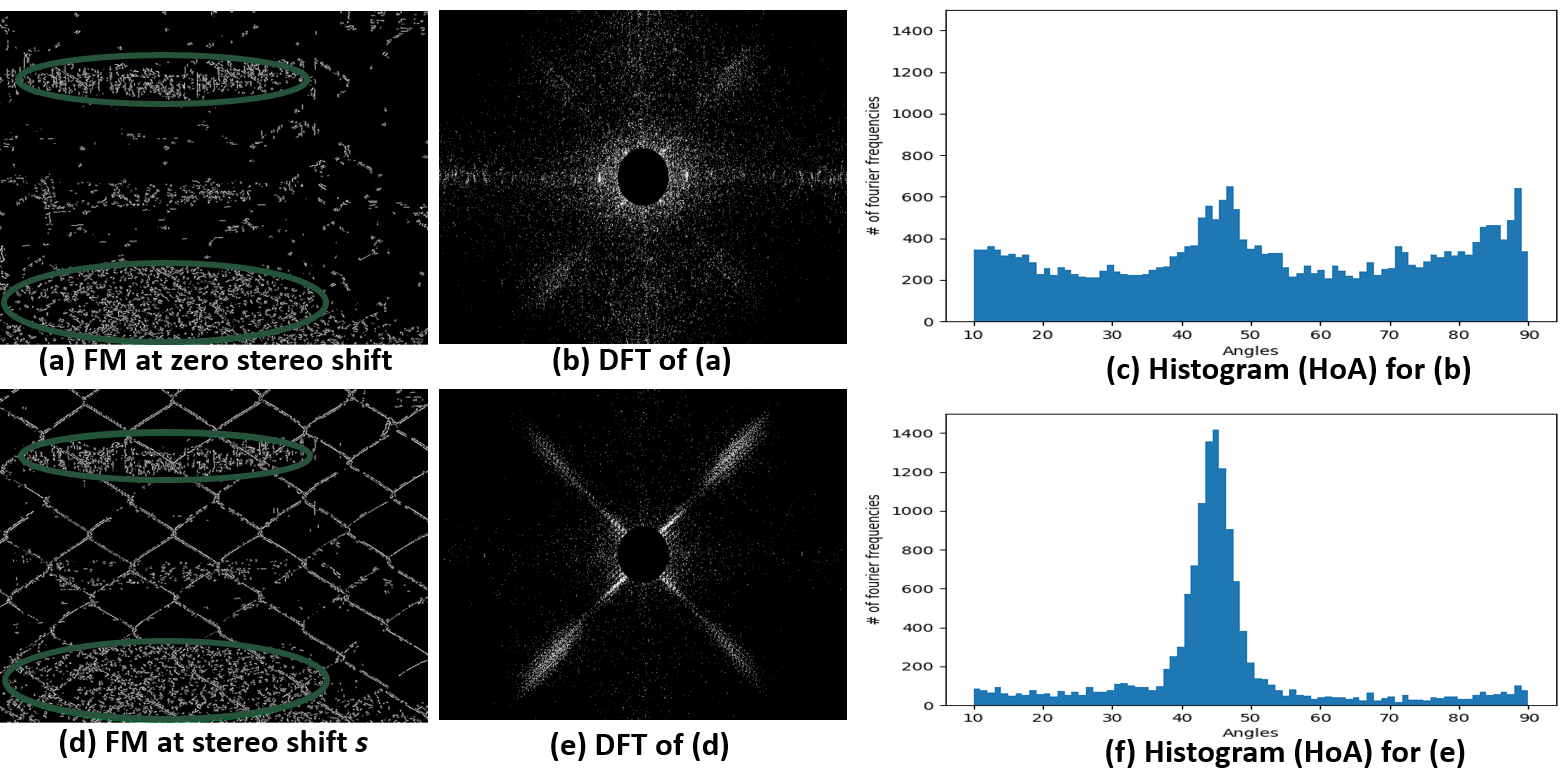}
	\caption{Fence guidance mask generation}
	\label{fig: 2}	
	\vspace{-1\baselineskip}
\end{figure}
$CL_i$ is the image $CL$ shifted by $i$ pixels and $FM_i$ is the subsequent fence mask for $i^{th}$ pixel shift. Fence edges appear in $FM_i$ for only a brief window of shifts when $i$ matches the foreground parallax ($s$) in pixels. Example of $FM_0$ and $FM_s$ in Fig. \ref{fig: 2} a \& d respectively illustrate this fact.  

To automate the estimation of precise pixel shift ($s$), we calculate $H_i$ (Fig. \ref{fig: 2} b \& e) as the Discrete Fourier Transform of $FM_i$. Since fence is a quasi periodic structure \cite{hettiarachchi2014fence}, its frequencies are aligned into lines or streaks in fourier domain (Fig \ref{fig: 2} e). These lines appear only when fence is visible in $FM_i$. We use this insight and calculate the Maximum Alignment Score ($MAS_i$) for every pixel shift $i$. The $MAS_i$ is defined in Eq. \ref{eq: 2} \& \ref{eq: 3} as the maximum number of frequencies having similar angle between their position ($x_p$,$y_p$) and image center ($x_0$,$y_0$).
\vspace{-0.25\baselineskip}
\begin{equation}
	\centering
	\label{eq: 2}
	\begin{split}
		Angles = \forall& _{x_{p},y_{p}} \text{  }array[\dfrac{180}{\pi} * \arctan(\dfrac{abs(y_p-y_0)}{abs(x_p-x_0)})]\\& \text{ $s.t.$ } 0<x_p<w;0<y_p<h\\
	\end{split}
\end{equation}
In Eq. \ref{eq: 2}, $w$ and $h$ signify width and height of $H_i$. The $Angles$ array contains all alignment measures between $0$ and $90$ degrees for fourier values in $H_i$. We eliminate the DC component of $H_i$ using a bandpass filter and consider values with $H_i[x_p,y_p] > \tau$ to eliminate small values. $\tau$ is empirically determined as $100$ for all images.  
\begin{equation}
\centering
\label{eq: 3}
\begin{split}
cHoA =& hist(Angles,numBuckets),\\ &\text{ $where$ }numBuckets=90\\
&MAS_{i} = max(cHoA)
\end{split}
\vspace{-0.25\baselineskip}
\end{equation}
In Eq. \ref{eq: 3} we divide the alignments from Eq. \ref{eq: 2} between $90$ buckets and then create a histogram of all angles (using $hist$). $hist$ returns the count of values ($cHoA$) for each bucket. $MAS_i$ is calculated as the maximum number of values aligned in one angle. When foreground pixel shift ($s$) matches $i$ we notice a sharp rise in $MAS_i$ (fig. \ref{fig: 2} c \& f). Hence, $s$ is computed using Eq. \ref{eq: 4} as the amount of pixel shift between stereo frames when $MAS_i$ is maximum.
\begin{equation}\label{eq: 4}
	\begin{gathered}
		s = i' \quad\text{ s.t }\quad MAS_{i'} \geq MAS_{i} \quad \forall i
	\end{gathered}
\end{equation}
Now using the precise pixel shift ($s$) and Eq. \ref{eq:1} we estimate the stereo-guided fence mask $FM_s$ or $FM$. Fig \ref{fig: 6} illustrates FM for multiple complex scenarios. % It may be possible to use the same method in scenarios when fence plane is not almost parallel with the camera plane. In such situations certain regions of $CL$ match at different $i$ with their stereo counterparts in $CR$. By dividing the stereo images into different regions and compute $s$ and $FM_s$ for each region separately. These $FM_s$ are stitched together to create the final guidance mask. 
\begin{figure}[b]
	\centering
	\includegraphics[width=0.1\linewidth,height=0.7cm]{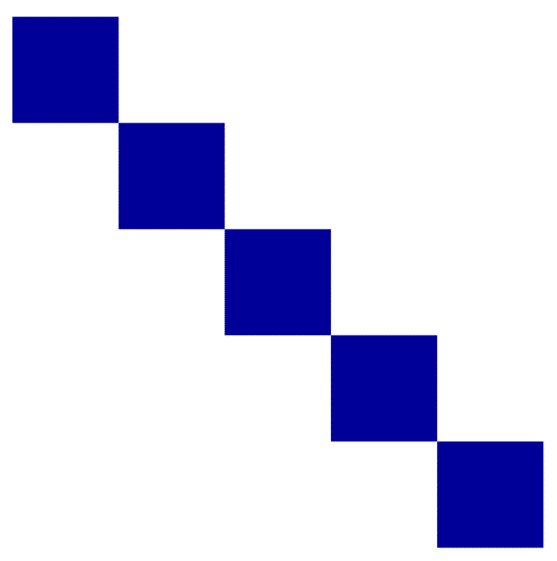}
	\includegraphics[width=0.1\linewidth,height=0.7cm]{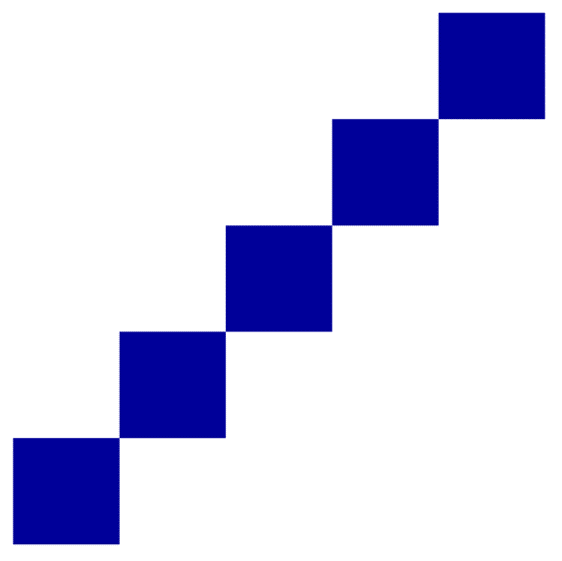}
	\includegraphics[width=0.1\linewidth,height=0.7cm]{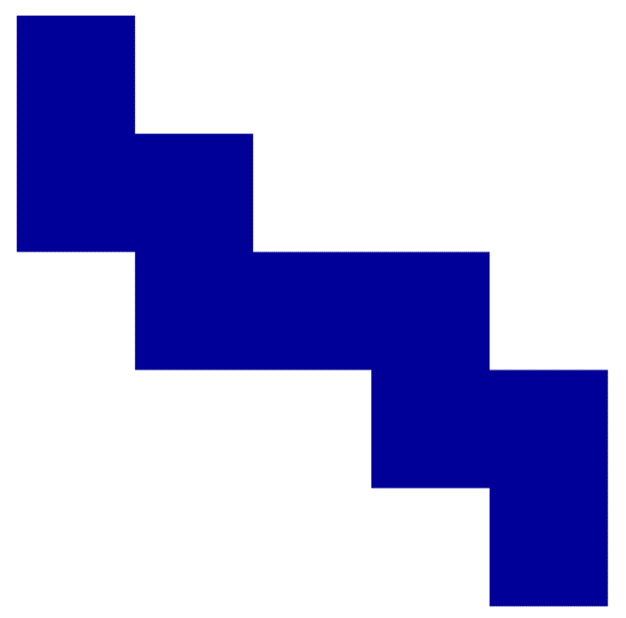}
	\includegraphics[width=0.1\linewidth,height=0.7cm]{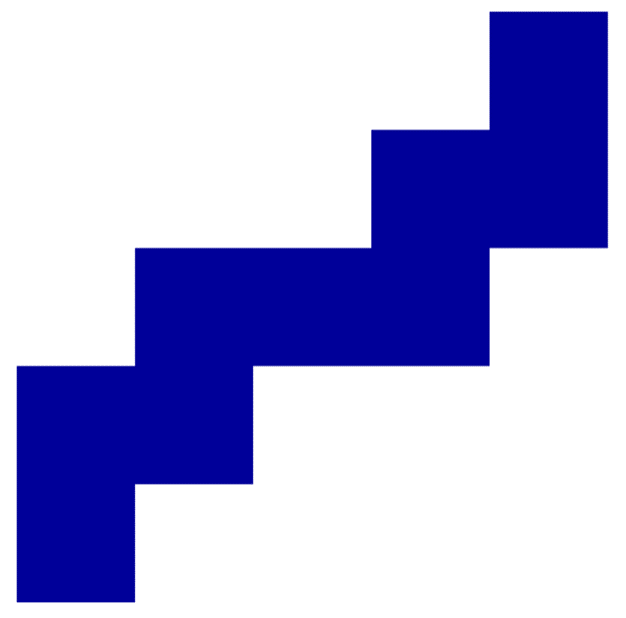}
	\includegraphics[width=0.1\linewidth,height=0.7cm]{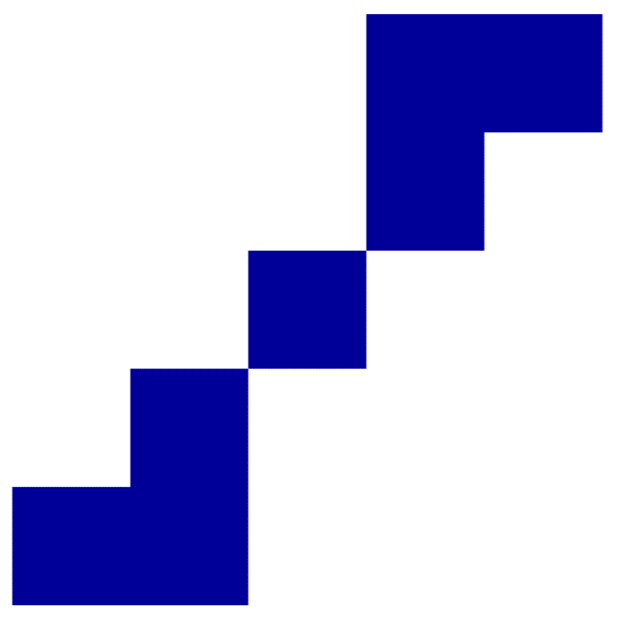}
	\includegraphics[width=0.1\linewidth,height=0.7cm]{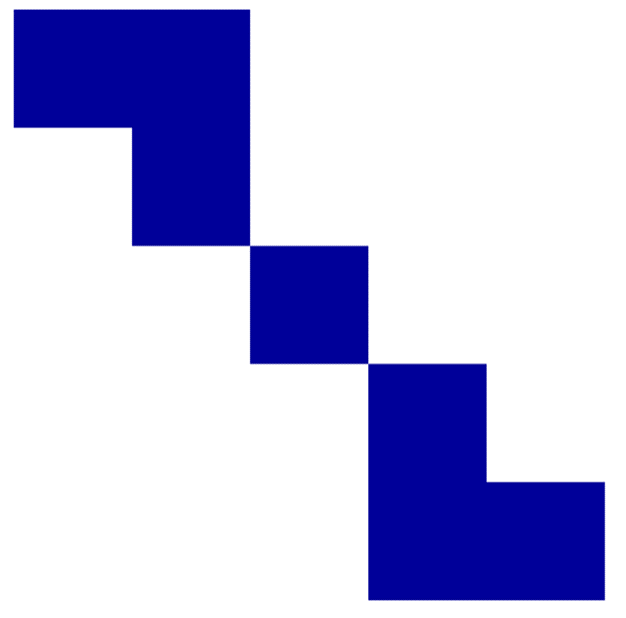}
	\includegraphics[width=0.1\linewidth,height=0.7cm]{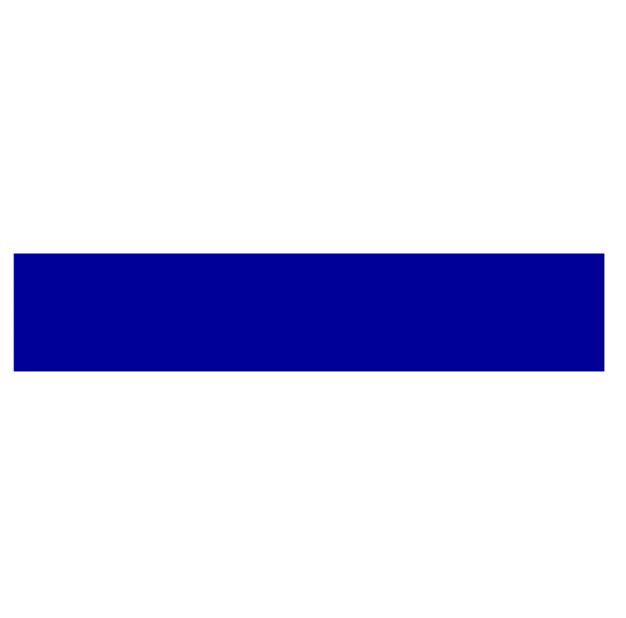}
	\includegraphics[width=0.1\linewidth,height=0.7cm]{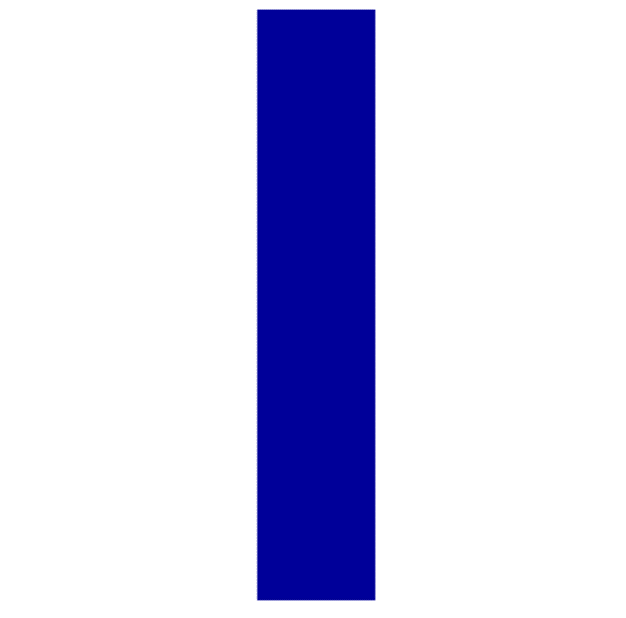}
	\caption{8 hand crafted features for DCL. Each is a 5X5 array where blue signifies 1 and white signifies 0.}
	\label{fig: 3}	
	\vspace{-0.5\baselineskip}
\end{figure}
\subsection{Deep fence estimation}
\label{subsec: fence model}
This work proposes two deep learning based fence estimation models, \textbf{DefenceGAN-3c} and \textbf{DefenceGAN-4c}. Both are Conditional Generative Adversarial Networks (GANs) and contain a Generator module (G) which aims to predict the fence template (y) from the given input image (x) and an adversary module (D), which validates the output of G. This additional deep neural network (D) results in an adversarial loss which dictates the generator network to create refined segmentation. In contrast to traditional per-pixel losses, an adversarial loss is generated from learned features. The Generator module is a U-Net like architecture with skip connections between encoder and decoder sub-modules. This ensures that both input and output are renditions of same underlying structure. DefenceGAN-3c is a single image based fence estimation model. DefenceGAN-4c uses both the single image and the generated guidance mask as input.% and furnishes superior results.% in particularly complex scenarios.% The flow of information from encoder to decoder through skip connections ensure that both input and output are renditions of same underlying structure. 
\begin{figure*}[t]
	\centering
	\includegraphics[width=0.9\linewidth,height=3.8cm]{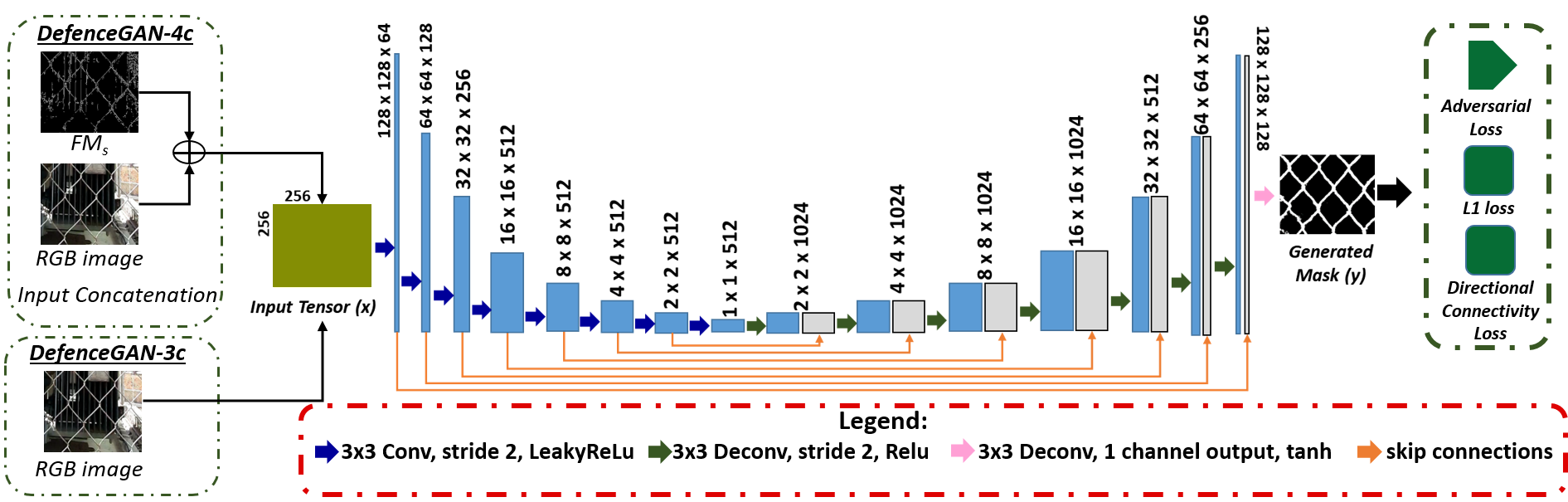}
	\caption{Architecture of proposed deep neural models}
	\label{fig: 4}	
	\vspace{-0.5\baselineskip}
\end{figure*}
\subsection{Directional connectivity loss}
\label{subsec: DCL-loss}
Fence joints are connected using thin wires which are generally difficult to detect. The predicted fence segmentation mask often contains broken fence lines. We counter this problem by proposing a novel Directional Connectivity Loss (DCL), used in combination with Adversarial Loss and L1-loss. The DCL (Eq.\ref{eq: 7}) is designed to enforce connectivity by penalizing pixels that are not connected with one another in accordance with $8$ hand crafted directional features (F) pictorially represented in Fig. \ref{fig: 3}. The proposed features cover almost all angle possibilities in a 5x5 neighborhood of pixels due to the discrete nature of pixels. These features are noteworthy in their similarity to the hand crafted Haar features from Viola \& Jones \cite{viola2001rapid}.
% It is generally difficult to segment these thin wires particularly in situations when fence is not in focus, or when background is too complex and rich in texture.
\begin{equation}\label{eq: 7}
	\begin{gathered}
		%DCL = \sqrt{(z\odot F[x,y,f_i]) - (y\odot F[x,y,f_i])^{2} + \epsilon}
		DCL = - \dfrac{1}{W*H} * \sum_{w \in W} \sum_{h \in H} \max_{f_i \in [1,8]}(y \odot F[w,h,f_i])
	\end{gathered}
\end{equation}

Here $W$ and $H$ are the width and height of final output ($y$). We examine the 5x5 neighborhood of every predicted pixel and compute a per-pixel connectivity score by convolving the generated fence mask (y) with these hand-crafted directional features. We train our proposed models to maximize this score (minimize DCL), thereby enforcing connections between fence wires.

%Considerable amount of accidental subtractions in training data ensure that model in not overfitted on subtracted guidance mask. 
%We find these pseudo-stereo image pairs sufficient for training. This is because the additional subtracted guidance mask from our artificial dataset is very similar to that from original stereo images.
% a U-Net like setup with skip connections between encoder and decoder sub-networks. This 
% We use 200,000 pairs as training samples and 25,000 pairs as testing samples.
% Below is an example of how to insert images. Delete the ``\vspace'' line,
% uncomment the preceding line ``\centerline...'' and replace ``imageX.ps''
% with a suitable PostScript file name.
% -------------------------------------------------------------------------

\section{Experimental Results}
\label{sec:Results}
Due to lack of publicly available large scale stereo fence images, our models are trained on an artificial dataset. We validate the robustness of our proposed approach by performing qualitative analysis on real world images and quantitative analysis on images from our dataset and from \cite{du2018accurate}.  Our models are trained using Adam optimizer \cite{kingma2014adam} with a batch size of 32 images and constant learning rate of $0.0002$.
\subsection{Data creation}
We create an artificial pseudo-stereo dataset of 250,000 image pairs split into 200,000 train and 50,000 test pairs. The diverse images from ImageNet \cite{imagenet_cvpr09} dataset are used as background. A mix of real fence regions from \cite{du2018accurate} and 97 hand-segmented fence templates from real videos are used as foreground. We randomly and unidirectionally shift both regions and ensure that foreground shift is greater than that of background. We also perform affine transformations, image cropping  and color distortions in images to create a diverse dataset. Random salt and pepper noise is added to the FM to imitate accidental subtractions as explained in section \ref{sec:propApp}.

\begin{figure*}[t]
	\centering
	\includegraphics[width=0.9\linewidth,height=9.8cm]{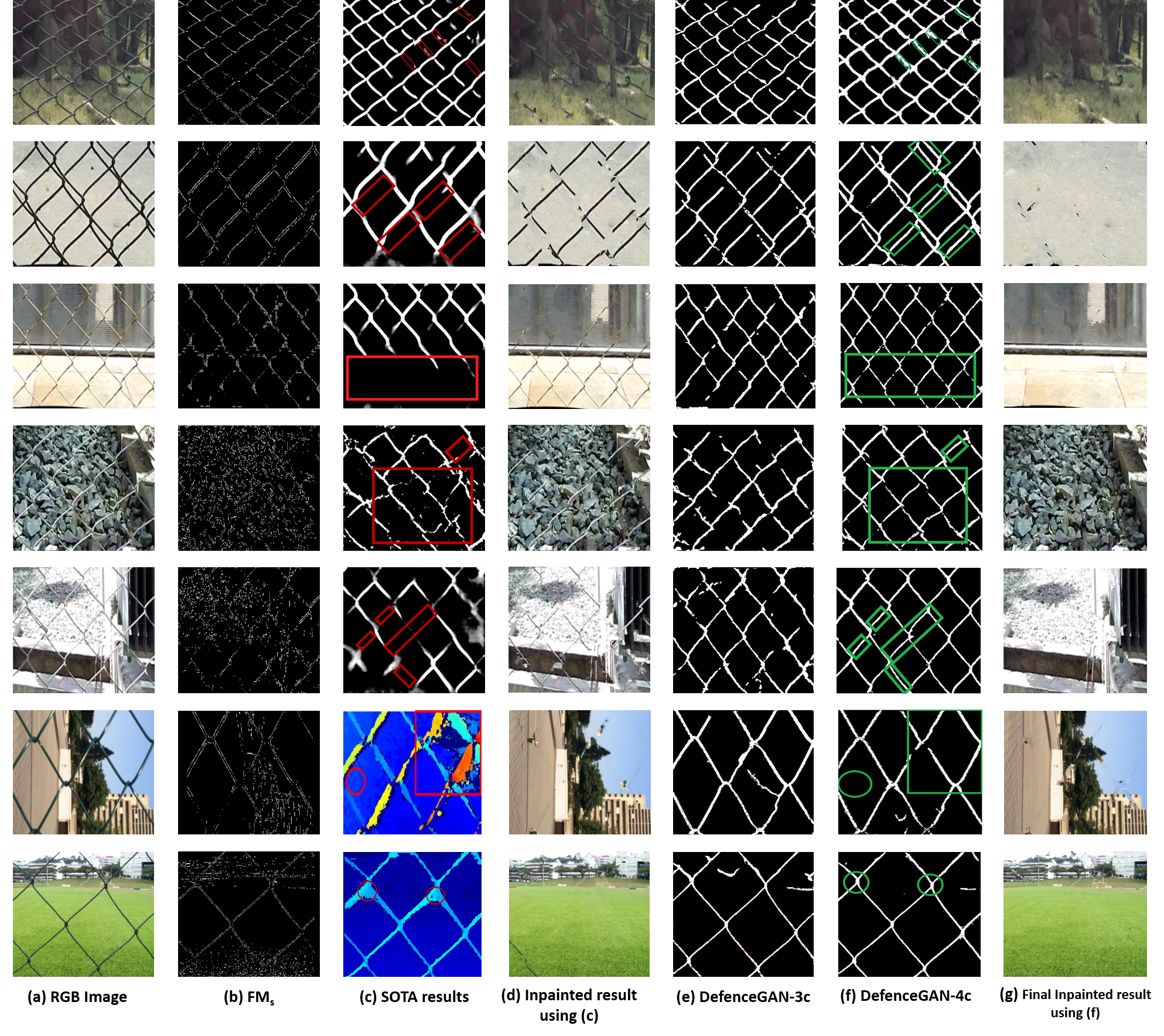}
	\caption{Row I to V showcase results on complex images from our real world test data. Row VI \& VII exhibit results on images from \cite{Mu2014VideoD,jonna2017stereo}}
	\label{fig: 6}	
	
\end{figure*}

Due to lack of depth information, the two images are not ideally stereo. However, the FM generated using artificial image pairs is visually indistinguishable to that created using real stereo pairs. Also, since the model uses single image and FM during both training and inference, it does not depend on stereo depth information. Fig.\ref{fig: 6} demonstrates that trained model performs well in real scenarios.
%It works well for real scenarios as illustrated in Fig. \ref{fig: 6}.
%We validate the robustness of our proposed approach by performing quantitative analysis (presented in Table 4$\&$5) on our test data and making qualitative comparisons (Fig. 6) with \cite{du2018accurate} and \cite{jonna2017stereo} on images from \cite{Mu2014VideoD,jonna2017stereo,yi2016automatic} and on real life complex scenes.\\
%For quantitative analysis, 
\subsection{Quantitative analysis}
We first compare results of single image based DefenceGAN-3c with Du \etal \cite{du2018accurate} and park \etal \cite{4815261} on all 100 test images in publicly available dataset from \cite{du2018accurate}. Since Du \etal \cite{du2018accurate} did not provide results without video based temporal refinement (TR), we replicate their refinement stage for effective comparison. Results in Table \ref{tab:quanti} illustrate the superiority of DefenceGAN-3c. To the best of our knowledge, there is no stereo image based fence estimation work performing quantitative analysis and the unavailability of public stereo image dataset containing fence as obstruction makes comparison with other works infeasible. Using transitivity, we prove the superiority of DeFenceGAN-4c by drawing quantitative comparisons with DefenceGAN-3c (Table \ref{tab:quantDefencGAN-4c}). We ensure consistency of our results by performing five-fold cross validation. Table \ref{tab:quantDefencGAN-4c} contains the mean ($\mu$) and standard deviation ($\sigma$) of all five sets. 
\begin{table}[t]
	\centering
	\caption{Quantitative Analysis on dataset from \cite{du2018accurate}}
	\vspace{-0.5\baselineskip}
	\label{tab:quanti}
	\resizebox{0.9\columnwidth}{0.12\linewidth}{%
		\begin{tabular}{|c|c|c|c|}
			\hline
			Method & Precision & Recall & F-Measure \\ \hline
			Park \etal \cite{4815261} & 0.500 & 0.163 & 0.246 \\ \hline
			Du \etal \cite{du2018accurate} with TR & 0.910 & 0.959 & 0.934 \\ \hline
			DefenceGAN-3c w/o TR & 0.947 & 0.872 & 0.908 \\ \hline
			\textbf{DefenceGAN-3c with TR} & \textbf{0.922} & \textbf{0.961} & \textbf{0.941} \\ \hline
		\end{tabular}
	}
	\vspace{-0.75\baselineskip}
\end{table}
\begin{table}[b]
	\centering
	\caption{Quantitative Analysis on our dataset}
	\vspace{-0.75\baselineskip}
	\label{tab:quantDefencGAN-4c}
	\resizebox{\columnwidth}{0.14\linewidth}{%
		\begin{tabular}{|c|c|c|c|c|c|c|}
			\hline
			\multirow{2}{*}{Method} & \multicolumn{2}{c|}{Precision} & \multicolumn{2}{c|}{Recall} & \multicolumn{2}{c|}{F-measure} \\ \cline{2-7} 
			& $\mu$ & $\sigma$ & $\mu$ & $\sigma$ & $\mu$ & $\sigma$ \\ \hline
			DefenceGAN-3c & 0.876 & 0.0088 & 0.809 & 0.0096 & 0.842 & 0.0079\\ \hline
			\begin{tabular}[c]{@{}c@{}}DefenceGAN-3c\\ with DCL\end{tabular} & 0.887 & 0.0069 & 0.828 & 0.0061 & 0.857 & 0.0060 \\ \hline
			DefenceGAN-4c & 0.960 & 0.0043 & 0.939 & 0.0042 & 0.949 & 0.0037\\ \hline
			\textbf{\begin{tabular}[c]{@{}c@{}}DefenceGAN-4c\\ with DCL\end{tabular}} & \textbf{0.973} & \textbf{0.0034} & \textbf{0.964} & \textbf{0.0027} & \textbf{0.968} & \textbf{0.0026} \\ \hline
		\end{tabular}
	}
\end{table}
\subsection{Qualitative analysis}
Fig. \ref{fig: 6} compares proposed DefenceGAN-3c and DefenceGAN-4c with works from Du \etal \cite{du2018accurate} (row I to V) and Jonna \etal \cite{jonna2017stereo} (row VI to VII). We use DCL to train both our proposed models. We do not take into account the improvements due to temporal refinement as proposed in \cite{du2018accurate} because it depends on capturing a video for every scene. We also fine-tune the model proposed by Du \etal \cite{du2018accurate} on single images from our dataset to ensure consistency. The results of DefenceGAN-4c (Column VI) are evidently superior when compared with outputs (column III) from current state-of-the-art (SOTA) methods. Certain regions in column III of fig. \ref{fig: 6} are demarcated in red to highlight the drawbacks of current best methods. Corresponding regions in column VI are demarcated in green to show the efficacy of our proposed method. 

Rows I to V in fig. \ref{fig: 6} contain real scenes that we captured. Row II is a complex double fence image very different from training data. The model trained without adversarial learning fails to adapt towards this variation. Row III, IV \& V have regions of similar background and foreground color, complex pebbles and strong sunlight respectively resulting in highly inaccurate single image based fence segmentation (Column III \& V). Using the fence guidance mask ($FM$) greatly enhances the segmentation ability of DefenceGAN-4c in these scenarios. Rows VI \& VII contain standard stereo images from \cite{yi2016automatic,Mu2014VideoD,jonna2017stereo} and are evaluated against the results by Jonna \etal \cite{jonna2017stereo}. We can observe that output from \cite{jonna2017stereo} contain regions of inaccurate disparities that are nicely predicted using DefenceGAN-4c. We also notice that background objects in these examples are very far away from fence and hence the computed disparity map is highly influenced by only fence pixels. This is generally not the case in many real life scenarios. The morphological transformations and matting on disparity map as used in \cite{jonna2017stereo} may not produce satisfactory fence segmentation. We overcome this limitation by leveraging the deep features extracted by DefenceGAN-4c to greatly enhance the correctness of predicted fence.

After fence segmentation, we apply existing single image inpainting technique \cite{Criminisi:2004:RFO:2319036.2320602} to remove fence from the image. Column VI contains the inpainted results using fence segmentation from SOTA methods (Column III) while Column VII contains the results of inpainting using predicted fence mask from proposed DefenceGAN-4c (Column VI). Considerable visual improvements in fence removal clearly establishes the importance of precise fence segmentation in the overall process. We believe better fence removal can be achieved by exploiting stereo or multiple frames.%, which is our future work.
%Note that row IV highlights an interesting ability of the proposed model to generate refined segmentation even with noisy guidance mask.
\subsection{Ablation Study}
We perform additional experiments (included in Table \ref{tab:quantDefencGAN-4c} and Fig.\ref{fig: 6}) to further analyze the role of DCL and guidance mask ($FM$) in fence segmentation. For DCL, we train similar instances of proposed models with and without this loss. Table \ref{tab:quantDefencGAN-4c} highlights an average increment of 0.011 in precision and 0.019 in recall for DefenceGAN-3c and an average increment of 0.013 in precision and 0.025 in recall for DefenceGAN-4c on our dataset.

We evaluate the impact of guidance mask ($FM$) by making quantitative and qualitative comparisons between DefenceGAN-3c and DefenceGAN-4c. Both models are trained identically with the latter having an addition inference time input ($FM$). Table \ref{tab:quantDefencGAN-4c} highlights a major improvement of \textbf{0.086} in average precision and \textbf{0.136} in average recall on our dataset when we use $FM$. This considerable improvement was also consistent across all five-folds of cross validation. Rows III, IV \& V in Fig. \ref{fig: 6} also demonstrate the enhanced ability of our proposed DefenceGAN-4c in real life complex scenarios mainly by the use of a novel inference time additional input. This indicates that our proposed use of $FM$ and DCL significantly enhances the segmentation of fence wires.
\section{Conclusion and Future Work}
\label{sec:Conclusion}
In this paper, we introduce a novel way of precisely segmenting fence in an image using a deep neural network with adversarial loss and a guidance mask generated from stereo image pairs. This mask contains the approximate shape of fence and hence enables the accurate prediction of fence regions in complex scenes using a single click in multi-camera setup. We introduce two neural networks that, for the first time, leverage adversarial learning for refined predictions and better generalization on training data. We also enforce a connectivity loss (DCL) to better connect fence joints especially when it is difficult to identify fence wires. We perform quantitative and qualitative analysis of our proposed method with current state-of-the-art fence estimation approaches and report superior results particularly in difficult situations.

\bibliographystyle{IEEEbib}
\bibliography{References}

\end{document}